\documentclass[sigconf,nonacm]{acmart}

\AtBeginDocument{%
  }

\setcopyright{cc}
\copyrightyear{2024}

\usepackage{enumitem}
\usepackage{graphicx}
\usepackage[framemethod=TikZ]{mdframed}
\newcommand\roundcorner{5pt}

\setlist[itemize]{leftmargin=1em}

\begin{document}

\title{User Simulation in the Era of Generative AI: User Modeling, Synthetic Data Generation, and System Evaluation}

\author{Krisztian Balog}
\affiliation{%
  \institution{University of Stavanger}
  \country{Norway}
}
\email{krisztian.balog@uis.no}

\author{ChengXiang Zhai}
\affiliation{%
  \institution{University of Illinois at Urbana-Champaign}
  \country{USA}
}
\email{czhai@illinois.edu}

\begin{abstract}
User simulation is an emerging interdisciplinary topic with multiple critical applications in the era of Generative AI.  It involves creating an intelligent agent that mimics the actions of a human user interacting with an AI system, enabling researchers to model and analyze user behaviour, generate synthetic data for training, and evaluate interactive AI systems in a controlled and reproducible manner. 
Because of its broad scope, research on this topic currently remains scattered across artificial intelligence, human-computer interaction, information science, computational social science, and psychology. To address this fragmented landscape of current research, this article presents a foundational synthesis. We highlight the paradigm shift from traditional predictive models to modern generative approaches, and explicitly frame critical ethical considerations---demonstrating how controlled simulation serves not merely as a risk vector for bias, but as a powerful, proactive tool to ensure fair representation and system safety. Furthermore, we establish the theoretical connection between user simulation and the pursuit of Artificial General Intelligence, arguing that realistic simulators are indispensable catalysts for overcoming critical data and evaluation bottlenecks and optimizing personalization. Ultimately, we propose a practical, self-sustaining innovation ecosystem bridging academia and industry to advance this increasingly important technology.
\end{abstract}


\maketitle

\section{Introduction}

Advancements in generative artificial intelligence (AI) are presenting unprecedented opportunities for innovation across diverse fields, while simultaneously introducing complex challenges.
Effective personalization in generative AI demands precise user modeling, which can be difficult to achieve due to the complexity and variability of individual preferences. Additionally, the success of interactive AI algorithms relies heavily on access to comprehensive interaction data for training, posing challenges in data collection and privacy. Finally, the evaluation of these AI systems is particularly difficult at scale via repeatable and reproducible experiments because of their interactive nature.
User simulation, which involves using an intelligent agent to mimic a real user's decisions during interactions with an AI system, offers a promising solution to all these challenges. By providing a controlled environment for training, testing, and refining AI systems, user simulation becomes a key enabler of safe and responsible advancements in generative AI. This article provides an overview of this important emerging topic, highlighting its interdisciplinary nature and its broad impact. 

As a research topic, user simulation is inherently interdisciplinary, intersecting with diverse fields both within and beyond computer science. For example, it draws upon concepts from psychology, economics, and human-computer interaction to create accurate and representative models of user behaviour~\citep{Balog:2024:FnTIR}. The recent success of large language models (LLMs) has made it possible to simulate complex user actions and led to their widespread use in simulation tasks across different domains and application scenarios, including
generating realistic conversations for dialogue systems~\citep{Kim:2021:ACL},
performing automatic relevance assessment of search results~\citep{Thomas:2024:SIGIR},
simulating specific human subpopulations in social science research~\citep{Argyle:2023:PA},
and simulating the behaviour of communities~\citep{Park:2023:UIST}. 
Realizing the potential of user simulation, there has been a surge of interest and activity in this area, evidenced by the growing number of related workshops~\citep{Balog:2021:SIGIRForum,Breuer:2024:SIGIRForum,Schaer:2025:SIGIRForum} and tutorials~\citep{Balog:2024:WWW,Deng:2024:WWW,Balog:2025:SIGIR,Zerhoudi:2026:CHIIR}.

This article aims to synthesize research dispersed across various fields, review the current state of the art, and highlight potential future directions. Rather than proposing new models or architectures, our primary contribution is to provide a foundational overview. By connecting fragmented efforts across disciplines, we aim to establish a shared conceptual framework.
Specifically, after outlining the paradigm shift from traditional predictive modeling to modern generative approaches, we provide a brief historical review of how simulation techniques can be employed for 
(1) user behaviour modeling, by creating realistic simulations of how users interact with a system when attempting to complete some task,
(2) data augmentation, by generating synthetic user interactions to improve the training of machine learning models, and
(3) system evaluation, by measuring the perceived utility and cost/effort from a user's perspective when completing a task.
Next, we broaden our perspective on user simulation to explore its role in integrating multiple research fields. Following this, we address the critical ethical considerations inherent in user simulation. Crucially, we reframe these challenges to demonstrate how simulation serves not merely as a risk vector for bias, but as a powerful, proactive tool to debias training pipelines, ensure fair subpopulation representation, and systematically stress-test safety guardrails. Finally, we discuss the role user simulation may play in the quest for Artificial General Intelligence (AGI), the ultimate goal of developing AI systems with human-like intelligence. To ensure the continuous and scalable advancement of this technology, we conclude by proposing a practical, self-sustaining innovation ecosystem that bridges academia and industry.

We believe that this synthesis will be valuable not only to researchers but also to engineers and practitioners working broadly with generative AI techniques.
Additionally, given the numerous difficult open challenges and the substantial work that remains, this article will also be of interest to funding agencies, as addressing these issues will require substantial support and investment.

\section{User Simulation}
\label{sec:usersim}

What exactly do we mean by user simulation? In essence, it involves creating an intelligent agent that mimics how a user interacts with a system. This agent can be built based on models/algorithms/rules and any knowledge we have about the user (their behaviour, knowledge, etc.).  Crucially, the agent can be parameterized to simulate a diverse range of users with varying characteristics. While historically constrained by narrow, task-specific action spaces, the advent of Generative AI has fundamentally transformed how these agents are constructed, a paradigm shift toward more holistic user simulation we explore in detail below.

\begin{figure}[t]
    \centering
    \includegraphics[width=0.5\textwidth]{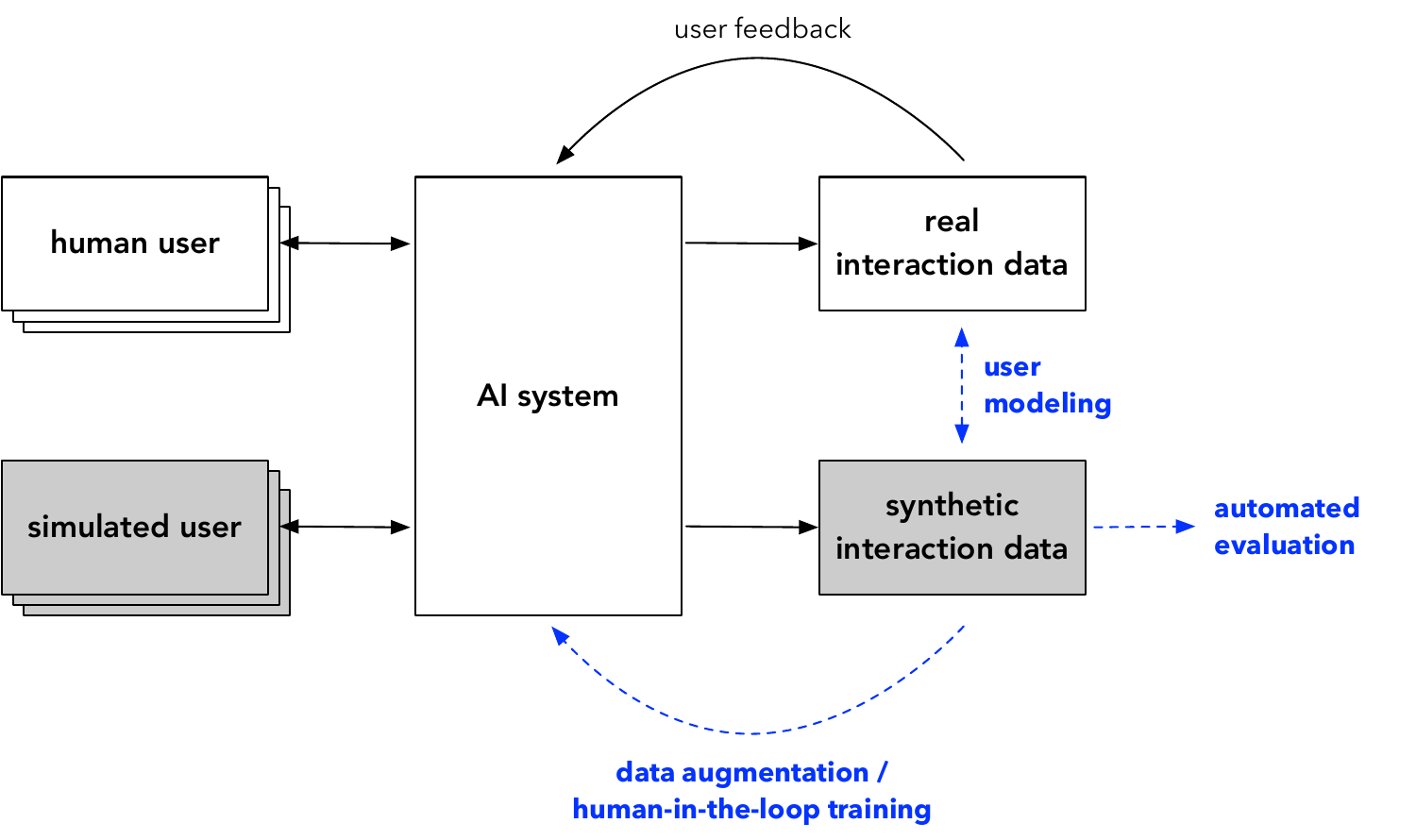} 
    \caption{Overview of the various uses of user simulation.}  
    \label{fig:overview}
\end{figure}

As illustrated in Figure~\ref{fig:overview}, once constructed, a user simulator can be used with any interactive AI system for evaluation or training purposes. Specifically, simulation has the potential to enable repeatable and reproducible evaluations at a low cost, without using invaluable user time (human assessor time or online experimentation bandwidth).  In addition, simulation can augment traditional evaluation methodologies by offering possibilities to gain insights into how system performance changes under different conditions and user behaviour.
Simulation techniques can also be leveraged to generate large amounts of synthetic data, which can be used for training AI models, especially in scenarios where real data is scarce or expensive to collect. Moreover, user simulation facilitates human-in-the-loop training, where human feedback is integrated into the learning process, known as Reinforcement Learning from AI Feedback (RLAIF)~\citep{Bai:2022:arXiv}. Finally, by comparing the observed real user interaction data with the synthetic data produced by an interpretable user simulator, we can model any user or any group of users and test hypotheses about their behaviour and preferences.  

\begin{table*}[t]
    \centering
    \normalsize
    \caption{Examples of user simulation, ranging from single actions to more complex behaviours.}
    \label{tbl:simulation_examples}
	\centering
	\begin{tabular}{p{4cm}p{4cm}p{4.7cm}p{3.1cm}}
		\toprule
		\textbf{Task context} & \textbf{System environment} & \textbf{User characteristics} & \textbf{Action space} \\
		\midrule
            Rating a product to express satisfaction 
            & E-commerce website with product pages and rating features 
            & User's purchase history, browsing behaviour, and demographic information 
            & Browsing, Rating \\
		\midrule
		Finding a movie to watch & Recommender system with slates of items &
		Previous watch history & Clicking, Watching \\
		\midrule
		Collecting information about a given topic 
            & Search engine with a query box and navigable search result lists 
            & University researcher conducting a comprehensive literature review on a topic 
            & Querying, Clicking \\
		\midrule
            Writing a complex piece of code for a specific functionality
            & AI-powered programming tool (IDE)
            & Previous code written or reviewed
            & Providing instructions, reviewing AI-generated code, making edits \\
		\midrule
            Learning a new skill
            & AI-powered tutoring system
            & Learning style, pace, prior knowledge
            & Completing exercises, asking questions \\
            \bottomrule
	\end{tabular}
\end{table*}

\subsection{Definition}

User simulation is the process of modeling a user's behaviour and decision-making patterns within an interactive system, specifically designed to mimic and predict how a user will act in various interaction contexts or scenarios related to completing a task.
To effectively simulate a user's behaviour within an interactive system, three core dimensions that influence this behaviour must be defined:
\begin{itemize}
	\item \emph{\textbf{Task context}}: A user's behaviour varies according to nature of the user's task. Tasks vary in complexity, and different tasks require different types and levels of interaction, decision-making processes, and completion strategies.
	\item \emph{\textbf{System environment}}: A user's behaviour depends on the system they interact with. This includes the system's functionality, user interface, and overall usability and support for task goals. It is the system that dictates the types of possible actions that a user can perform at any given point during their interactions. 
	\item \emph{\textbf{User characteristics}}: Different users may behave differently when completing the same task using the same system. Simulations must account for variations in individual user characteristics such as age, technical proficiency, preferences, and cognitive styles.
\end{itemize}
With these dimensions defined, the task of user simulation can be stated as the following computational problem:
\begin{quote}
    \emph{The goal of user simulation is to create an agent that can simulate every action a specific user may take when attempting to complete a specific task using a specific system.}
\end{quote}
This problem involves developing an intelligent agent (i.e., a user simulation agent) that can dynamically generate user actions, reflecting the behavioural patterns and decision-making processes of a user, based on a specific system environment and task context. 
The main technical challenge is to compute a decision-making policy.  
This policy must evaluate the current state---encompassing information about the task, the user's traits, the system environment, and the history of previous interactions---to determine the next most plausible action. Once an action is taken, the environment transitions to a new state, leading to a new decision problem. While the choice of computational model for the policy (e.g., rule-based, probabilistic, or machine-learned algorithm) varies, such user simulation agents can generally be modeled based on the Markov Decison Process (MDP) framework (see~\citep{Balog:2024:FnTIR} for a deeper discussion on instantiating existing approaches within the MDP framework).

\subsection{Scope}

User simulation encompasses a wide spectrum, ranging from predicting single actions to modeling complex behaviour across multiple tasks. In our formulation, this scope is primarily determined by how the task context is defined. At one end of the spectrum, this context might represent a very specific interaction, such as predicting whether a user would click on a particular search result snippet. Here, the focus is on simulating a single, isolated action. Moving along the spectrum, the context could encompass a sequence of actions within a given session, such as reformulating search queries, requiring the model to consider dependencies between actions. Further expanding the scope, it might represent an entire overarching task, such as finding information on a particular topic or completing a purchase, where the simulation would involve multiple sequences of interactions. Finally, at the broadest level, the scope could encompass a user's general preferences and behaviour across various tasks, necessitating models that capture long-term patterns and adapt to different environments. Thus, by varying the granularity and breadth of the task context, our formulation allows for user simulations in a wide range of application scenarios at different levels of complexity.
Table~\ref{tbl:simulation_examples} lists specific examples of user simulation for tasks.

\subsection{Traditional Approaches}

The problem of simulating the action a user takes in a given context (represented by the decision policy) can often be framed as a classification problem when there is a relatively small set of actions to choose from; for example, simulation of a user's clicking action may be framed as a binary classification problem, where the algorithm would predict whether the simulated user would click on a specific item. With the problem framed as a classification problem, different approaches generally vary in how they perform the classification (equivalently prediction) task. 
At a high level, we can distinguish two broad approaches: model-based and data-driven.

\begin{itemize}
	\item \textbf{\emph{Model-based}} approaches may be based on rules designed with knowledge about how users behave or on interpretable probablistic models that can more flexibly capture uncertainties using interpretable parameters.
	The parameters of such models may be set heuristically or empirically derived from observed user data. By varying those parameters, different types of users can be simulated.
	\item \emph{\textbf{Data-driven} (or machine-learned)} approaches emphasize maximizing accuracy of fitting any observed real user data, without necessarily imposing interpretability. Almost all such approaches are based on supervised machine learning, notably using deep neural networks which can learn effective, but non-interpretable representations from the data for predictive modeling.
\end{itemize}
\noindent
These two families of approaches may also be combined, e.g., by utilizing model-based techniques to compute effective features for data-driven approaches or employing machine-learned models in specific components of model-based approaches.

\begin{table*}[t]
    \centering
    \normalsize
    \caption{Comparison of traditional user simulation approaches versus the Generative AI paradigm.}
    \label{tbl:approaches_comparison}
    \begin{tabular}{p{2.2cm}p{4.5cm}p{4.5cm}p{4.7cm}}
        \toprule        
        & \textbf{Model-based} & \textbf{Data-driven} & \textbf{Generative AI} \\
        \midrule
        \textbf{Core \mbox{mechanism}}
        & Explicit rules or interpretable probabilistic models
        & Supervised machine learning (e.g., deep neural networks) 
        & Foundation models (LLMs) pre-trained on broad world knowledge \\
        \midrule
        \textbf{Action space} 
        & Highly constrained (predictive selection from predefined choices)        
        & Constrained to specific patterns present in the training data 
        & Open-ended (natural language, code generation, complex sequential actions) \\
        \midrule
        \textbf{Primary strengths}
        & High interpretability; allows precise counterfactual tuning and parameter variation 
        & High predictive validity; effectively captures complex implicit patterns directly from interaction logs 
        & Zero-shot generalization; capable of complex, multi-turn interactions; easily parameterizable via prompting \\
        \midrule
        \textbf{Limitations \& weaknesses} 
        & Rigid; heavily relies on manual feature engineering; struggles to capture complex, nuanced human behaviour 
        & ``Black box'' nature with low interpretability; requires large amounts of interaction data; lacks zero-shot generalization 
        & Prone to behavioural hallucination; difficult to strictly control cognitive constraints (e.g., bounded rationality) \\ 
        \bottomrule
    \end{tabular}
\end{table*}

When there are potentially infinitely many actions to choose from (e.g., when formulating a query, any valid query would be potentially an option), simulating user actions becomes significantly more challenging for traditional approaches. The vast action space can lead to computational complexity and makes it difficult to define a meaningful probability distribution over possible actions. In practice, we often make assumptions to restrict the number of actions to be considered when simulating those actions (e.g., assuming the user will only use keywords from a predefined vocabulary, or that the query will follow a specific structure).
However, modern approaches, notably the large generative models driving the current wave of Generative AI, can handle large or infinite action spaces much more effectively.

\subsection{The Generative AI Paradigm Shift}

Building on this capability to handle open-ended action spaces, the advent of Generative AI introduces a broader technical paradigm shift: the transition from \emph{predictive} to \emph{generative} simulation. Historically, user simulation predominantly relied on predictive classification---such as predicting whether a user will click a specific link or assign a specific rating. These traditional models require task-specific feature engineering, struggle to generalize beyond their training data, and are entirely incapable of handling open-ended interactions. In contrast, rather than learning isolated behaviours from scratch, LLMs bring foundational world knowledge, semantic understanding, and zero-shot generalization to the simulation environment. This allows GenAI-based simulators to operate within infinite, unconstrained action spaces---such as formulating natural language queries, engaging in complex multi-turn dialogues, or generating novel code---which were previously impossible to simulate at scale. Consequently, GenAI elevates user simulation from a narrow, task-specific evaluation tool into a versatile, foundational building block capable of modeling complex cognitive tasks.
To summarize this evolution, Table~\ref{tbl:approaches_comparison} provides a comprehensive comparison of traditional model-based and data-driven approaches against the emerging Generative AI paradigm, highlighting their respective core mechanisms, action spaces, strengths, and limitations.

\subsection{Uses of Simulation}

User simulation has many uses, including:
\begin{itemize}
	\item Gaining insight into user behaviour to inform the design of systems and evaluation measures.
	\item Performing large-scale automatic evaluation of interactive systems (i.e., without the involvement of real users), and analyzing system performance under various conditions and user behaviours (answering \emph{what-if} questions, such as ``What is the influence of X on Y?'').
	\item Augmenting data with human feedback and generating synthetic data with the purpose of training machine learning models and addressing data scarcity or privacy concerns. More broadly, user simulation can facilitate machine learning approaches that require human input (interactive learning, reinforcement learning, or human-in-the-loop systems).
\end{itemize}
We will review these major applications (user modeling, data augmentation, and system evaluation) in detail later, but here in this section, we first discuss the desirable properties of a user simulator and the trade-offs between multiple dimensions. 

\subsection{Requirements and Desiderata}

\begin{table*}[t]
    \centering
    \normalsize
    \caption{Desirable properties of user simulators and their inherent trade-offs and challenges.}
    \label{tbl:desiderata}
    \begin{tabular}{p{3.3cm}p{6cm}p{6.9cm}}
        \toprule
        \textbf{Property} & \textbf{Definition} & \textbf{Key trade-offs \& challenges} \\
        \midrule
        \textbf{Validity} 
        & Simulated behaviours align with empirical observations of real user behaviour in similar contexts 
        & High predictive validity often comes at the cost of interpretability, as highly accurate data-driven models are frequently opaque \\
        \midrule
        \textbf{Interpretability} 
        & The simulated behaviour can be understood and adjusted through explicitly controllable parameters 
        & Often limits the complexity and predictive power of the model; highly interpretable models may fail to capture nuanced user interactions \\
        \midrule
        \textbf{Cognitive plausibility} 
        & Decision-making processes are grounded in theories or models of human cognition, preventing arbitrary actions 
        & Translating complex psychological theories into computable algorithms is fundamentally difficult; current LLMs mimic idealized reasoning rather than actual human cognitive constraints (e.g., bounded rationality, fatigue, and limited memory) \\
        \midrule
        \textbf{Variation} 
        & Incorporates natural human variability and occasional outliers rather than merely replicating average behaviour 
        & Difficult to calibrate effectively; excessive variation can easily lead to arbitrary, unrealistic, or unsafe simulated behaviours \\
        \midrule
        \textbf{Adaptability} 
        & User simulator agents can learn from their interactions, updating their expectations and adjusting behaviour accordingly over time 
        & Requires complex, dynamic memory mechanisms (e.g., maintaining evolving knowledge states) and significantly increases overall simulation complexity \\
        \bottomrule
    \end{tabular}
\end{table*}
 
Designing an effective user simulator requires balancing several complex, and often competing, requirements. Table~\ref{tbl:desiderata} summarizes the core desirable properties of user simulators, outlining their definitions alongside the inherent trade-offs and challenges involved in achieving them.
\begin{itemize}
    \item \textbf{Validity}: Simulated users must exhibit behaviours that align with empirical observations of real user behaviour in similar contexts. This includes both high-level strategies (e.g., information seeking patterns) and low-level actions (e.g., clicking behaviour). Without validity, the insights gained from simulation cannot be trusted. 
    \item \textbf{Interpretability}: While not strictly a requirement, interpretability is a highly desirable property. Interpretability means that the simulated behaviour can be understood and adjusted through controllable parameters. This allows researchers to (1) understand why the simulator produced certain behaviours, (2) investigate how changes in specific parameters influence the behaviour of users, and (3) systematically introduce controlled variation into the simulated population. Since user behaviour and preferences vary significantly across users, interpretability is crucial for understanding which real-world users can be expected to produce results similar to the simulations.
    \item \textbf{Cognitive plausibility}: The decision-making processes underlying simulated user behaviour should be grounded in theories or models of human cognition, ensuring that the simulated actions are not arbitrary or random. A cognitively plausible simulator can be expected to generalize better. 
    \item \textbf{Variation}: Simulated users should reflect general user behaviour patterns but also incorporate variability and occasional outliers, rather than merely replicating average behaviour.  That is, simulation should reflect the unpredictable nature of real human interactions. 
    \item \textbf{Adaptability}: Simulated users should be able to learn from their interactions with the system, update their expectations, and adjust their behaviour accordingly. This includes adapting to changes in their own knowledge, preferences, and interaction patterns over time.
\end{itemize}
While it is generally desirable to optimize a user simulator across all those dimensions, we often have to prioritize certain dimensions over others.
For example, while both validity and interpretability are desirable, there often exists a trade-off between the two. Data-driven (machine-learned) simulators, trained on large datasets of real user behaviour, can often achieve high predictive accuracy, capturing complex patterns and nuances in user actions. However, this predictive power comes at the cost of reduced interpretability. The internal workings of these models, often involving complex neural networks or ensemble methods, can be opaque and difficult to understand. This makes it challenging to pinpoint the specific reasons behind a simulated user's behaviour or to adjust the model's parameters in a controlled manner. 

The good news is that simulation does not need to be perfect in order to be useful in a specific application context. 
Note that evaluation methodologies, even in well-established areas like search engine ranking, are inherently imperfect. For example, relevance judgments in almost all test collections are incomplete. However, research suggests that the relative performance of different retrieval systems tend not be affected much by this incompleteness~\citep{Voorhees:2000:IPM}.
In fact, creating a ``perfect'' user simulator, i.e., one that flawlessly replicates human behaviour across all possible tasks and contexts, is likely an AI-complete problem, on par with achieving Artificial General Intelligence (AGI).

\section{User Simulation for User Modeling} 
\label{sec:usermodeling}

Understanding users is key to any personalized system, and the the study of user behaviour in interactive systems has been the subject of decades of user-oriented research~\citep{Kelly:2009:FnTIR}.
Conceptually, a user model is an explicit representation of the user's background, goals, and expected behaviour when interacting with a system.
User models can be utilized to optimize the design of user interfaces and algorithms, enabling them to adapt effectively to the specific needs of each individual user. 

Analysis of user behaviour is often performed via controlled user studies.
However, a major challenge in user studies is that the experimental conditions can introduce variables that are difficult to control (e.g., users' background, domain knowledge, or perseverance), but they can affect how users interact with the system.
Another approach to studying user behaviour is to analyze historical log data generated by real users.
While log-based studies allow for the verification of certain assumptions about user behaviour, they are limited to a specific version of the system in use at the time the logs were collected. As a result, they do not support testing hypotheses that involve changes to the system, such as evaluating the effects of varying response quality or different response times.

From the perspective of user modeling, user simulation can be regarded as taking a computational approach to developing a comprehensive and functional user model.
It requires the creation of a mathematical model of the user, serving as a formal hypothesis about their behaviour.
This hypothesis can be rigorously tested for validity through user studies or historical logs, by comparing how closely the behaviour of simulated users match that of real users. 
Such comparisons can also provide valuable insights into improving the simulator itself. 
Thus, advancements in user simulation directly contribute to a deeper understanding of real-world users.

Moreover, simulation allows for the examination of the potential effects of system modifications on anticipated user behaviour, offering a more comprehensive assessment than a live environment typically permits.
This includes, for instance, estimating long-term engagement and satisfaction across diverse user populations that can be controlled or varied in a counterfactual manner. 
These insight-gaining type of simulations can answer questions like: What has the greatest influence on X? or How will X and Y interact?
Simulation-based studies have led to many useful findings about users that suggest ways to improve systems. For example, it has been demonstrated that users find items more effectively when recommendations are presented as list of carousels instead of a simple ranked list~\citep{Rahdari:2024:TORS}.

Lastly, as a valuable tool for user studies, a formal user simulator can facilitate various comparative analyses in association with any relevant user variables. For instance, by partitioning a large user population along various dimensions (e.g., age, occupation, or geo-locations), user simulation models can be employed to conduct comparative analyses among different subgroups to understand their commonalities and differences as reflected in the observed user data such as interaction logs. Similarly, any variable related to an AI system can also be varied to examine how a user's behaviour may vary depending on system-related variables; for example, the device and time available can profoundly affect what kind of interactive searching behaviour can be successful~\citep{Baskaya:2012:SIGIR}.

With a simulator of an individual user as a building block, it is also possible to simulate a community of people, which, theoretically speaking, can be any set of individuals that are connected in some way.  
By representing the behaviour of a community as an aggregate function of the simulated behaviours of its individual users, we can study its overall dynamics and evolution over time.
For example, a community may include the entire user base of an application such as a recommender system, and user simulation can be used to study the long-term impact of, for instance, how content creators' strategic behaviour in competing for user engagement~\citep{Yao:2023:ICML} or the tendency of users choosing more popular items~\citep{Hazrati:2024:UMUAI} influences the overall content ecosystem and user experience within the platform. 
Modeling a community introduces the complex challenge of simulating how users would interact with each other---an area LLMs will likely play an important role~\citep{Park:2023:UIST}. 

\section{User Simulation for Data Augmentation} 
\label{sec:dataaug}

As data is the ``fuel" to power AI systems, both its quality and quantity directly affect system effectiveness. 
The type of data available also determines the kind of intelligence a system can acquire. For example, the abundance of textual data on the Web has enabled the training of powerful LLMs like ChatGPT and Gemini, which have acquired linguistic knowledge and much common sense knowledge. However, user interaction data is scarce, as generating it requires actual user interactions, which are time-consuming and expensive to collect.
For instance, even the search log data of a major commercial search engine would still be limited in its coverage of diverse information needs, user behaviours, and preferences. 
Moreover, user interaction data generally contains sensitive user information, restricting its access for researchers outside the company where it is collected.  User simulation offers a natural solution by generating large amounts of synthetic user interaction data. This can be done in various ways, including augmenting existing user interaction datasets for offline model training or employing user simulators in interactive, human-in-the-loop training scenarios. 

While interpretability is crucial for user simulation in evaluation, it is less critical for data augmentation. The usefulness of synthetic data often depends more on its distributional similarity to real user data than on the simulator's realism. 
For instance, in Reinforcement Learning from AI Feedback (RLAIF)~\citep{Bai:2022:arXiv}, the learned reward model, which implicitly captures user preferences, can be considered a non-interpretable user simulator.
This relaxed requirement---that interpretability is not essential for effective data augmentation---coupled with advances in deep learning, particularly the rise of LLMs capable of generating realistic text, has facilitated progress in user simulation for data augmentation, especially in the context of conversational systems~\citep{Soudani:2024:arXiv}. However, concerns remain about the diversity of LLM-generated text~\citep{Chung:2023:ACL}.  Interpretable simulators, with their ability to vary meaningful user parameters in a counterfactual way (e.g., by making a simulated user increasingly more patient), offer greater control and flexibility for data augmentation compared to non-interpretable ones.

It is worth noting that while simulating users holistically is desirable, even simulating specific user actions can still be valuable for generating additional task-specific training data (e.g., generating queries that might lead to clicks on a particular item~\citep{Dai:2023:ICLR}). Focusing on specific actions would also simplify making progress in data augmentation. 
Since data augmentation benefits various machine learning models, utilizing user simulation for this purpose expands its scope into potentially many other areas of AI research.

\section{User Simulation for Evaluating Interactive AI Systems} 
\label{sec:eval}

Evaluation of AI systems is crucial for both advancing AI research and deploying production systems. In research, rigorous evaluation allows us to compare algorithms and understand the relative strengths and weaknesses of different approaches. This knowledge guides further research and development. For deployed systems, evaluation ensures that the chosen algorithm meets the specific needs and expectations of its users.  Inaccurate evaluation can misdirect research efforts or lead to the deployment of suboptimal systems, hindering progress and potentially having negative consequences for users.

There are three widely-used evaluation methodologies for AI systems: reusable test collections, user studies, and online evaluation.

\textbf{Reusable test collections} (or offline evaluation) enable large-scale automatic evaluation, facilitating algorithm comparison and advancement. They ensure repeatability and allow for the study of individual components within complex methods. However, these static evaluations often fail to capture user interactions and behaviours adequately, relying on simplified models of processes and user behaviour. This methodology, while standard for making relative comparisons between systems, struggles with evaluating the actual utility of interactive systems due to its inherent abstraction and deviation from the dynamic complexities of real-world applications.

\textbf{User studies} capture real users' interactions in controlled settings, providing high-fidelity insights into a system's actual utility. However, they are costly, time-consuming, and often suffer from reproducibility issues due to user fatigue and learning effects. Moreover, recruiting a sufficient and representative user base often proves challenging, making this method less accessible for smaller companies and academic settings.

\textbf{Online evaluation} (or log-based studies) observes real users interacting with a fully operational system, offering reliable measurements of quality and user experience through large-scale data analysis. A/B testing is a common example~\citep{Kohavi:2020:book}. While this method enables utility assessment with many users, it lacks control over user variables, complicating result interpretation. Additionally, it cannot accommodate counterfactual evaluation and poses risks if new system versions perform poorly, potentially impacting user perception of the system.

\begin{figure*}[t]
    \centering
    \includegraphics[width=0.65\textwidth]{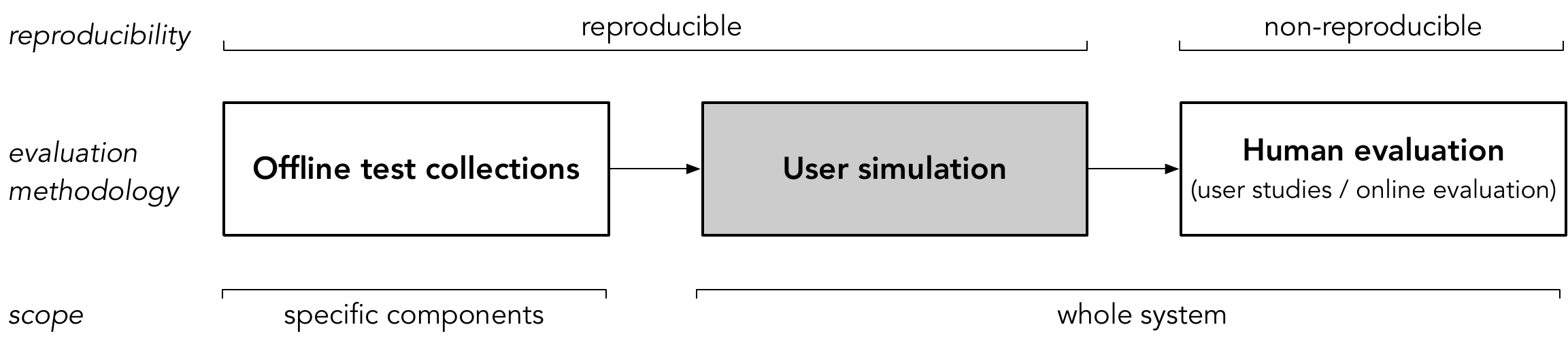} 
    \caption{Illustration of evaluation methodologies and how user simulation complements them (adapted from~\citep{Balog:2024:FnTIR}).}  
    \label{fig:evaluation}
\end{figure*}

In general, all the three methodologies above can be applied to, and indeed have been regularly used for, evaluating interactive AI systems. 
However, these approaches  have limitations when it comes to conducting reproducible experiments: the test collection-based approach is static by its very nature, while there is an inherent lack of reproducibility when real users are involved. User simulation can help address these challenges: simulated users can be controlled, enabling reproducible experiments and systematic analysis of system performance. Figure~\ref{fig:evaluation} illustrates how user simulation relates to and complements these traditional evaluation methodologies.

Generally, we may consider interactions with an AI system as ``moves'' in an interactive game, where the user and the AI system take turns making moves based on the current interaction environment. 
A user simulator is an operational agent that can simulate how a user makes those decisions.
To evaluate an AI system, we can let it interact with a simulated user and measure its performance based on the interaction data generated. This approach allows for a controlled and systematic evaluation of how the system responds to different user behaviours and preferences.

A general simulation-based evaluation framework can thus be defined as consisting of the following elements:
\begin{enumerate}
    \item A collection of user simulators are constructed to approximate real users.
    \item  A collection of task simulators are constructed to approximate real tasks.
    \item  Both user simulators and task simulators can be parameterized to enable modeling of variation in users and tasks.
    \item  Evaluation of a system is done by first having a simulated user perform a simulated task by using (interacting with) the system and then computing various measures based on the entire interaction history of the whole ``task session.''
\end{enumerate}
From a user's perspective, we can measure the performance in two dimensions:
(1) Interaction Reward,  which measures the total reward the user has received via the interaction, and (2) Interaction Cost, which measures the total cost of the interaction. In general, the more interaction actions the user makes, the more reward the user can potentially receive and the more cost the user would have to bear (since the user needs to make more effort). If one single measure is needed, the reward and cost can be combined, which can be in many different forms, all reflecting the intuition that an ideal system should maximize reward to a user while minimizing the cost or effort required of the user. With such a simulation-based evaluation framework, many specific measures can be defined.
As discussed in~\cite{Balog:2024:FnTIR}, the existing test collection-based evaluation methodology can be viewed as a simple form of user simulation. 
This means user simulation is actually already being widely utilized especially in the context of information access tasks, albeit implicitly, without explicitly articulating what kind of users are being simulated. 
A key advantage of evaluation based on user simulation is that
assumptions about simulated users and their behaviour are made more explicit. User simulation also enables us to evaluate {\em interactive} AI systems that would otherwise be hard to evaluate, especially if we want experiments to be reproducible (e.g., evaluating a dynamically generated user interface for browsing~\cite{Zhang:2017:ICTIR}). 
For a comprehensive treatment of simulation-based evaluation of interactive systems, we refer the reader to the recent book by \citet{Balog:2024:FnTIR}.

\section{User Simulation as an Interdisciplinary Research Field}
\label{sec:broader}

User simulation research is inherently interdisciplinary, intersecting with a multitude of fields, including that of information science, information access, machine learning, natural language processing, knowledge representation, human-computer interaction, and psychology.
Recognizing this, future research will benefit significantly from integrating insights and approaches from these diverse areas.
Below, we briefly discuss some of the connections and potential synergies between user simulation and related fields.

\subsection{Intelligent Agents}
A sophisticated user simulator can be regarded as an intelligent agent interacting within a system's environment. Consequently, techniques used for developing intelligent agents can also prove valuable in the construction of user simulators. More specifically, multi-agent systems are particularly well-suited for simulating communities of users, as they enable the modeling of complex interactions among multiple individuals within the simulated environment~\citep{Uhrmacher:2009:book}. A perfect user simulation agent can also be regarded as a basis to implement a human digital twin, which has widespread applications~\cite{Wang:2024:RCIM}.
Conversely, building upon user simulation research, intelligent agents can benefit from even more precise modeling of individual behaviour.

\subsection{Machine Learning}
The extent to which machine learning is incorporated into user simulators can vary based on interpretability needs. For applications demanding high interpretability, like user modeling, machine learning might be employed for specific simulator components. When interpretability is less critical, as in data augmentation, end-to-end simulators can even be directly learned from observational data.
From the perspective of machine learning, there are numerous tasks that require human involvement, such as labeling, annotation, human-in-the-loop learning, and model evaluation.
User simulation offers a promising avenue to streamline these processes by reducing reliance on human labor, thereby cutting costs and saving time. Notably, recent research exploring the use of LLMs for data labeling and relevance assessment can be viewed as developing specialized user simulators that emulate the actions of human annotators~\citep{Thomas:2024:SIGIR,Chiang:2023:ACL}.
Another pertinent example is reinforcement learning with human feedback (RLHF), which has recently gained significant attention for its use in training LLMs~\citep{Ouyang:2022:NeurIPS}. 
In RLHF, a machine-learned reward model is trained to capture human preferences and subsequently used to guide the training of the primary machine learning model---this reward model is essentially a user simulator.
    
\subsection{Knowledge Representation}
A realistic user simulator requires a model of the user's knowledge state, encompassing both their general world knowledge and specific understanding of the application domain and the AI system itself. As user actions are often a consequence of their particular knowledge state, user simulation is inherently linked to the study of knowledge representation and involves reasoning over knowledge (to decide which action to take).
Furthermore, the user's knowledge model needs to be dynamically updated to reflect learning through interaction with the AI system~\citep{Vakkari:2016:JIS}.
While existing work has often relied on simple knowledge representations, future research could explore more sophisticated techniques, such as personal knowledge graphs~\citep{Skjaeveland:2024:AIOpen}, to enhance the realism of user simulators.
Broadly speaking, how to represent and reason over a user's knowledge, and how to capture their knowledge acquisition during interaction, remain crucial yet challenging open questions. Addressing these effectively may necessitate a deeper understanding of a fundamental AI question: how to formally represent knowledge.

\subsection{HCI and Psychology}
Psychology and HCI research are fundamental for building theoretically sound and realistic user simulators. 
Specifically, this can include insights gained from user studies, as well as well-motivated and explicit computational cognitive models that explain certain behaviours, such as navigation behaviour on the Web~\citep{Fu:2007:HCI}, which can directly inform the design of interpretable user simulators.
At the same time, an interpretable user simulator can be instrumental in analyzing user interaction data for deeper behavioural understanding.
Existing research has employed various models to analyze interaction logs; for example, mining search engine logs reveals benefits of users following search trails versus jumping directly to destination pages~\citep{White:2010:SIGIR}.
Furthermore, simulators can serve as testable ``computational models'' of users, enabling the formulation and validation of hypotheses about user behaviour. 

\subsection{Software Systems}
User simulators need to function as operational software platforms. 
Given that sophisticated user simulators will likely increase in complexity, efficiency challenges arise. 
This necessitates the use of advanced algorithms and data structures (e.g., for knowledge representation), distributed techniques to support large-scale experiments, and, in general, sound engineering practices.
User simulation offers numerous benefits for the development of interactive systems.
By simulating a variety of user behaviours and interactions, developers can gain insights into how their systems will perform in the real world, identify flaws before deployment, and make improvements accordingly~\citep{Bernard:2024:CUI}. 
This can lead to more robust, user-friendly, and efficient interactive systems, while also potentially reducing the reliance on human testers, thus saving time and resources.

\section{Ethical Considerations: Bias, Fairness, and Safety}
\label{sec:ethical}

As user simulation becomes increasingly integral to modeling, data augmentation, and system evaluation, it is critical to proactively address the ethical implications inherent in mimicking human behaviour. If left unchecked, biases embedded within a user simulator will inevitably propagate into the synthetic data it generates, creating a feedback loop that degrades the safety and fairness of any downstream AI models trained on that data. However, while user simulation introduces potential ethical risks, it concurrently offers a powerful mechanism to address these exact challenges in broader AI development. Rather than merely reflecting existing inequities, controlled simulation serves as a proactive tool to reduce bias, guarantee fair representation, and systematically stress-test system safety. Table~\ref{tbl:ethics_summary} summarizes these core ethical dimensions, contrasting the inherent risks with the proactive opportunities simulation affords.

\begin{table*}[t]
    \centering
    \normalsize
    \caption{Summary of the ethical dimensions in user simulation, contrasting inherent risks with proactive mitigation opportunities.}
    \label{tbl:ethics_summary}
    \begin{tabular}{p{2cm}p{6.5cm}p{7.5cm}}
        \toprule
        \textbf{Ethical} & \textbf{Inherent risk} & \textbf{Proactive opportunity} \\
        \textbf{Dimension} & \textbf{(if unmitigated)} & \textbf{(simulation as a tool)} \\
        \midrule
        \textbf{Bias \& data skew}
        & Simulators inherit and amplify societal biases or interaction skews present in uncorrected training corpora or historical logs 
        & Enables explicit data debiasing; simulators can generate balanced synthetic datasets to correct historical gaps before downstream training \\
        \midrule
        \textbf{Fairness \& representation}
        & Relying on a monolithic, ``average'' simulated user marginalizes minority groups and obscures disparate system performance 
        & Facilitates disaggregated evaluation; persona-based simulation ensures rigorous testing against underrepresented groups \\
        \midrule
        \textbf{Safety \& guardrails}
        & Simulators lacking proper constraints may inadvertently generate toxic, harmful, or unsafe behaviours during system interaction
        & Unlocks scalable red-teaming; adversarial user simulation intentionally stress-tests system vulnerabilities and safety policies without risking human harm \\
        \bottomrule
    \end{tabular}
\end{table*}

\subsection{Bias and Data Debiasing}
Generative foundation models are trained on vast, unfiltered internet corpora, meaning they inherently capture and reproduce historical societal biases~\citep{Bender:2021:FACCT} and often disproportionately reflect the opinions, values, and ideologies of specific dominant demographic groups rather than a globally representative population~\citep{Santurkar:2023:ICML}. Similarly, data-driven simulators inherit the interaction biases present in the logged historical data they learn from.  When these biased models are used to power a user simulator, there is a significant risk that the simulated interactions will skew toward stereotypes, over-represent dominant demographics, or provide skewed feedback to the system being evaluated. However, precisely because a simulator is a controllable agent, it can be utilized as a tool for bias mitigation. 
For example, early approaches to debiasing simulators demonstrated that correcting logged historical data prior to simulation effectively prevents existing interaction biases from degrading downstream policy optimization~\citep{Huang:2020:RecSys}. Furthermore, in scenarios where real-world interaction logs are heavily skewed (e.g., containing primarily data from younger, tech-savvy demographics), a user simulator can be parameterized to artificially generate balanced synthetic datasets. By generating interactions that intentionally compensate for historical data gaps, developers can debias the training pipelines of downstream AI systems, ensuring they learn from a more equitable distribution of user behaviours.

\subsection{Fairness and Diverse Representation}
Algorithmic fairness requires that an AI system performs equitably across all user demographics, regardless of language, culture, or socioeconomic status. A common pitfall in system evaluation is relying on a monolithic, ``average'' simulated user, which effectively marginalizes minority groups and obscure failure modes that only affect specific subpopulations. To counteract this, researchers can leverage persona-based user simulation. These approaches span a broad spectrum: ranging from prompting models with specific demographic variables~\citep{Argyle:2023:PA} to conditioning them on richly detailed narrative backstories that capture more nuanced, intersectional identities~\citep{Moon:2024:EMNLP}. By deliberately instantiating an array of simulated users that reflect diverse cultural backgrounds and varying levels of domain expertise, developers can conduct rigorous, disaggregated evaluations. This ensures that the AI system is explicitly tested against underrepresented or vulnerable user groups, allowing developers to identify and rectify performance disparities before real-world deployment.

\subsection{Safety and Adversarial Simulation}
The deployment of interactive AI systems carries inherent safety risks, ranging from the generation of toxic content to susceptibility to prompt injection attacks. If a user simulator lacks proper guardrails, it may inadvertently generate harmful behaviours during the training phase. Conversely, this capacity for negative behaviour can be harnessed intentionally for ``red teaming''---the practice of using AI to safely attack and test other AI systems~\citep{Perez:2022:EMNLP}. By configuring a user simulator to act as an adversarial user (e.g., one attempting to bypass safety filters, extract private information, or induce system errors), developers can expose edge cases and vulnerabilities at scale. In this context, user simulation becomes an indispensable security tool, allowing researchers to stress-test the boundaries of an AI system's safety policies in a secure, simulated environment without exposing real users to harm.
\section{User Simulation as a Step toward AGI}
\label{sec:agi}

The overarching goal of developing a realistic user simulator is, in many respects, fundamentally aligned with the broader objective of creating intelligent agents with human-like intelligence, namely, the pursuit of AGI.
A critical bottleneck hindering faster progress toward AGI is the heavy reliance on human interaction data for training and evaluation. This process is inherently slow, expensive, and difficult to scale. User simulation addresses this directly: by generating synthetic interaction data and providing scalable evaluation environments, it acts not merely as a useful tool, but as a critical catalyst required to accelerate AGI development~\citep{Balog:2026:CACM}.
Furthermore, this alignment is reflected in shared technological foundations.
Importantly, this is not a recent trend, but an empirical reality evident throughout AI history. From early rule-based expert systems to probabilistic models, and now to the current wave of Transformer architectures and LLMs, advancements in core AI technologies have consistently been leveraged simultaneously to build both more capable task agents and more realistic user simulators. Today, this structural alignment is most evident in how both user simulation agents and intelligent task agents share a common mathematical modeling approach, enabling optimization through reinforcement learning algorithms. 
The key differences lie in the definitions of states, actions, and reward functions. For user simulation, the goal (reward) is to mimic real user behaviour, while for task agents, the reward is typically tied to providing an effective and user-friendly service for task completion. Because of this deep synergy, the technical challenges in building intelligent user simulators mirror those encountered in developing intelligent task agents. Consequently, we anticipate that research into user simulation technology and intelligent task agents must advance hand-in-hand. The necessity of this co-evolution becomes particularly evident when examining the architectural requirements for true human-AI collaboration.

\begin{figure*}[!t]
    \centering
    \includegraphics[width=0.65\linewidth]{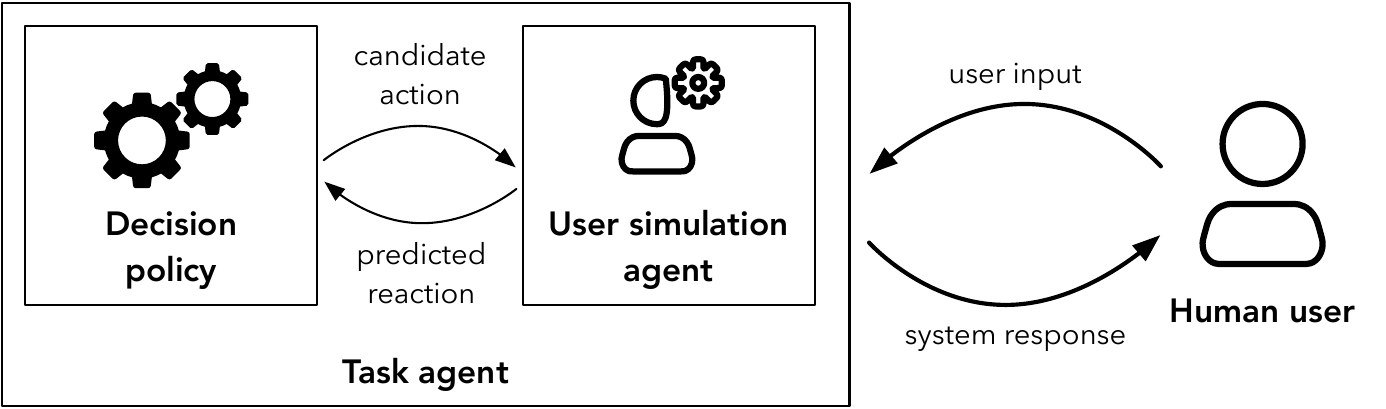}
    \caption{Conceptual ``Theory of Mind'' architecture for an advanced interactive AI system. The task agent utilizes an embedded user simulation agent as an internal world model. This creates a cognitive feedback loop where the decision policy proposes candidate actions, the user simulator forecasts the human's likely reaction, and the policy optimizes its choice before executing a final action in the real environment.}
    \label{fig:theory_of_mind}
\end{figure*}

\subsection{The Importance of User Simulation in Human-AI Collaboration}

Realizing the full potential of human-AI collaboration requires more than just superhuman performance from AI agents. To truly maximize combined intelligence, these agents must also account for the inherent variability in human behaviour, including suboptimal actions, diverse problem-solving approaches, and individual preferences.
Recent work in chess, a domain that has long served as a Petri dish for AI research, underscores this challenge. Studies have shown that human players paired with AI agents tailored to their skill level outperform those partnered with more powerful AI agents that are not adjusted for skill-compatibility~\citep{Hamade:2024:ICLR}.
This suggests that simply providing the most powerful AI assistance may not be optimal. Instead, simulating the user's knowledge, preferences, and decision-making processes, and adjusting the AI's assistance accordingly can lead to more effective collaboration.\footnote{Note that due to legal responsibility, even autonomous AI systems must collaborate with human owners to minimize risk and ensure safety. This requires the AI to simulate the human owner's value system and trade-off preferences.}
Such an approach requires user simulation agents to be closely integrated with task agents, allowing the task agent to leverage the user simulation agent to receive feedback and optimize its interaction policy. 
In essence, an autonomous intelligent agent that leverages its own internal user simulation agent to anticipate human needs and optimize interactions is implementing a functional analogue of Theory of Mind~\citep{Premack:1978:BBS}.
Figure~\ref{fig:theory_of_mind} illustrates this conceptual architecture. Rather than relying on purely reactive heuristics, an advanced task agent embeds a user simulation agent directly within its cognitive framework to act as an internal world model~\citep{LeCun:2022:OpenReview,Ding:2026:ACMSurvey}. During interaction, the task agent's decision policy does not immediately output a response to the real human user. Instead, it initiates an internal feedback loop: the policy proposes a candidate action, and the user simulator forecasts the human's likely response. By mentally ``rehearsing'' these interactions through simulated trial-and-error, the system can optimize its planning phase, ultimately executing actions in the real environment that are significantly more aligned, empathetic, and effective. 
Likewise, the user simulation agent must adapt to model how new AI systems behave over time. Therefore, the interdependence between research on intelligent task agents and user simulation is inherent and may persist until AGI is achieved.
For a more comprehensive discussion on the indispensable role of simulation in AGI, we refer the reader to~\citep{Balog:2026:CACM}.

\subsection{Large Language Models as Building Blocks for Intelligent Agents}

The emergence of LLMs could accelerate this integration and synergy, as these models may very well serve as a foundational building block for both types of agents.
Indeed, LLMs have already fueled their extensive adoption as both task agents and simulation tools in various domains and applications. 
For comprehensive surveys of work combining agent-based modeling and LLMs across various fields (social science, natural science, and engineering) and domains (e.g., cyber, physical, social environment) we refer readers to~\citep{Gao:2024:HSSC} and \citep{Wang:2024:FCS}. 

As the uses of LLMs for building user simulation agents and task agents are rapidly expanding, it is also crucial to recognize their limitations. 
LLM-generated responses can be unpredictable and sometimes unsafe, causing concerns about the safety of an intelligent task agent when deployed in the real world and the realisticity of a user simulation agent. For user simulation, current LLMs would inevitably exhibit unrealistic or incoherent behaviours and may also lack the natural variation observed in real human interactions. Furthermore, LLMs often possess more knowledge than average humans and generate overly ``perfect'' responses, which may not be a problem for an intelligent task agent (indeed, it could be a feature!), but would likely lead to the simulation of unrealistic ``superusers'' by a user simulation agent. 
While prompting techniques can guide LLM behaviour, ensuring strict adherence to instructions remains a challenge~\citep{Balog:2024:FnTIR}.
Importantly, more fundamental challenges suggest that LLMs alone will not be sufficient. Critically, LLMs have insufficient knowledge of human behaviour. While they might be aware of concepts like patience or satisfaction, they lack the training data to model the human dynamics of such behaviours. Similarly, LLMs lack a deep understanding of human cognitive processes, such as decision-making, memory recall, and attention span. They may fail to accurately simulate phenomena like cognitive biases or the limitations of working memory.  These shortcomings hinder their ability to generate realistic simulations of human users. 

To overcome this limitation, LLMs must be extended with components that capture a wider range of human cognitive abilities. The human brain is known to consist of two distinct systems: System 1, which is intuitive, fast, but not always reliable, and System 2, which is logical, deliberate, and slower~\citep{Daniel:2013:Book}. Current LLM-powered agents can simulate the rapid intuition of System 1 relatively well, but simulating System 2---and achieving seamless integration between the two---remains a formidable open challenge. For example, capturing uniquely human behaviours that require deep, deliberate reasoning, such as our capacity to not only use existing tools but invent new ones to solve unprecedented problems, remains a significant hurdle for current architectures~\citep{Xu:2025:DSE}.

\begin{figure*}[!t]
    \centering
    \includegraphics[width=0.65\linewidth]{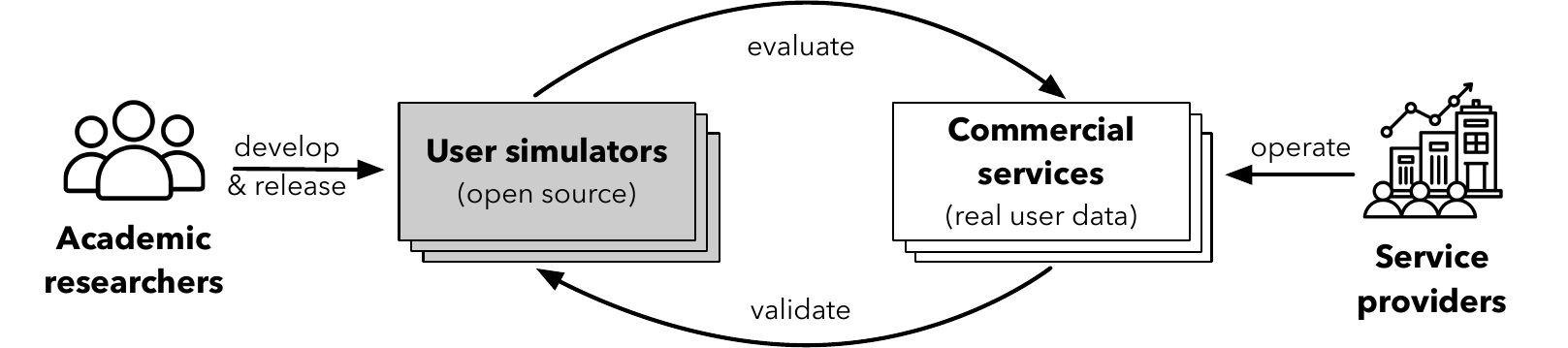}
    \caption{An innovation ecosystem, where academic researchers develop open-source user simulators, which industry partners validate using real user data, thereby bridging the data divide between academia and industry.}
    \label{fig:ecosystem}
\end{figure*}

Recent research has attempted to address these gaps through various architectural enhancements for LLM-powered systems. For instance, Retrieval-Augmented Generation (RAG)~\citep{Lewis:2020:NeurIPS} and GraphRAG~\citep{Edge:2024:arXiv} are increasingly used to ground an agent's responses in external knowledge, with specific retrieval tasks proving important for supporting human-like AI agents~\citep{Zhai:2025:SIGIR}. Furthermore, advanced prompting paradigms and agentic workflows---such as Chain-of-Thought (CoT)~\citep{Wei:2022:NeurIPS}, Tree-of-Thoughts (ToT)~\citep{Yao:2023:NeurIPS}, or the Reflection Pattern~\citep{Madaan:2023:NeurIPS}---can be employed to elicit multi-step reasoning and dynamically correct LLM errors.  Broader integrations of LLMs with cognitive architectures have also been explored to enable models to solve complex, open problems~\citep{Sumers:2024:TMLR}.

However, it is crucial to recognize the fundamental limitations of these techniques in the context of \emph{user simulation}. While RAG, CoT/ToT, and agentic reflection make LLMs better problem solvers, they do not necessarily make them more cognitively plausible. For instance, while a ``Critic'' agent is highly effective for optimizing a task agent's performance, applying such a cycle of self-criticism to user simulation is only valid if it is explicitly mapped to a human cognitive process, such as metacognition or conscious self-correction. Human reasoning is heavily constrained by bounded rationality, working memory limits, and cognitive biases. Therefore, simply prompting an LLM to ``think harder'' or ``critique itself'' to achieve optimal results does not adequately mimic the reality of human cognitive processes. Instead, achieving true cognitive plausibility requires the seamless integration of System 2 functionality into LLM architectures. Key challenges include determining how symbolic reasoning can be used to explicitly regulate agent behavior (e.g., via reward supervision), and how agents can acquire symbolic knowledge dynamically through interactions with their environment.  
In this direction, neurosymbolic approaches, which combine neural networks with symbolic reasoning, represent a much more promising future direction to capture the actual mechanisms of human thought~\citep{Garcez:2023:AIReview}. 

\section{Conclusion and Outlook} 
\label{sec:concl}

User simulation plays an increasingly important role in the era of Generative AI with  a wide range of applications, including user behaviour modeling and analysis, data augmentation, and system evaluation. 
Each application area may require different simulation techniques and will likely lead to the development of specialized simulators. At the same time, there is clear synergy among the different lines of exploration and the various research directions can enhance and inform one another. Over time, these diverse lines of research may even converge as we advance toward AGI. As this interdisciplinary community matures, a critical next step will be the development of unified metamodels, shared architectures, and exchange standards for user simulation. Currently, the lack of such standards hinders interoperability between simulators developed in different subfields. Establishing these frameworks will be essential for addressing the highly complex, multi-angle requirements of user simulation at scale. Much of the research today on user simulation is scattered across different communities. We envision a convergence of research in this area in the future, where researchers from multiple communities and disciplines would have increasingly more engagement and collaboration. The user simulation portal at \url{http://usersim.ai} is an emerging platform with the potential to facilitate this, though  further effort is required to accelerate interdisciplinary research and foster cross-disciplinary collaboration. In the short term, progress can be made by regularly hosting workshops on user simulation at different major conferences of different communities, which would facilitate the creation and growth of an interdisciplinary research community around the general topic of user simulation. 

\subsection{Establishing a Sustainable Innovation Ecosystem}

To facilitate validation of user simulators at scale, it is essential to involve industry partners who have access to real user behaviour data. 
Such collaboration can form the basis of a positively reinforced \emph{innovation ecosystem}, as illustrated in Figure~\ref{fig:ecosystem}. Rather than a one-sided transfer of technology, this ecosystem is designed so that all stakeholders act simultaneously as beneficiaries and contributors.
While academic researchers drive innovation by developing and open-sourcing novel user simulators, we acknowledge that expecting industry partners to validate these models without clear economic incentives is idealistic. Validating academic models against proprietary data incurs significant costs, including compute resources, engineering overhead, and strict data privacy compliance.
To offset these real-world constraints and ensure the ecosystem is self-sustaining, there must be a clear economic value exchange. With the proposed ecosystem, industry partners may be incentivized to bear these validation costs because the collaboration provides them with vital pre-competitive assets: independent benchmarks to prove their AI systems are safe, standardized evaluation frameworks they do not have to build from scratch, and direct engagement with an academic talent pipeline. Once validated against real-world data, the best simulators become directly useful for evaluating the industry's own product systems.
Meanwhile, this industry-provided validation service would attract academic researchers to the platform to test new ideas and algorithms. This continuous influx of academic innovation improves the overall quality of user simulators, which in turn reinforces the value proposition for industry partners, closing the loop on a self-sustaining collaboration platform.

As a practical first step toward this broader vision, the community can establish a scaled-down ``minimum viable ecosystem.'' Rather than beginning with complex, large-scale infrastructural commitments, initial progress can be driven through industry-sponsored shared tasks and public data challenges at major AI conferences. A prominent real-world example of this momentum is the newly launched User Simulation track at the Text REtrieval Conference (TREC),\footnote{\url{https://trec.usersim.ai}} organized by the National Institute of Standards and Technology (NIST). In models like this, an organizing body or industry partner releases a carefully anonymized, limited-scope interaction dataset, and academic teams compete to build the most accurate or cognitively plausible simulator. This approach requires significantly lower upfront investment, navigates legal and data privacy concerns more easily, and builds the foundational trust and baseline metrics necessary to validate the ecosystem concept. 

Eventually, to scale beyond limited datasets and fully mitigate data privacy constraints on a continuous basis, this validation can transition toward federated evaluation paradigms. Here, the open-source simulator is brought to the company's secure data environment, and only aggregated, anonymized validation metrics are made publicly available. Supported financially by joint industry consortiums or government grants, these shared validation results facilitate the assessment of progress, creating a positively reinforced and economically viable innovation loop between academia and industry.

\balance
\bibliographystyle{ACM-Reference-Format}
\bibliography{references}

\end{document}